\documentclass{article}
\usepackage{spconf}
\usepackage{amsmath,amssymb,graphicx,hyperref}
\usepackage{multirow}
\usepackage{booktabs}
\ninept

\title{LyTimeT: Towards Robust and Interpretable State-Variable Discovery}

\name{
Kuai Yu$^{1}$, 
Crystal Su$^{1}$, 
Xiang Liu$^{2}$, 
Judah Goldfeder$^{1}$, 
Mingyuan Shao$^{1}$, 
Hod Lipson$^{1,3}$
}

\address{
$^{1}$Department of Computer Science, Columbia University, New York, NY, USA\\
$^{2}$School of Computing, National University of Singapore, Singapore\\
$^{3}$Department of Mechanical Engineering, Columbia University, New York, NY, USA\\
\{ky2589, ys3791, jag2396, ms6592, hod.lipson\}@columbia.edu, \; liu.xiang@u.nus.edu
}

\begin{document}
\maketitle

\begin{abstract}
Extracting the true dynamical variables of a system from high-dimensional video is challenging due to distracting visual factors such as background motion, occlusions, and texture changes. We propose \textbf{LyTimeT}, a two-phase framework for \emph{interpretable variable extraction} that learns robust and stable latent representations of dynamical systems. In Phase~1, LyTimeT employs a spatio-temporal TimeSformer-based autoencoder that uses global attention to focus on dynamically relevant regions while suppressing nuisance variation, enabling distraction-robust latent state learning and accurate long-horizon video prediction. In Phase~2, we probe the learned latent space, select the most physically meaningful dimensions using linear correlation analysis, and refine the transition dynamics with a Lyapunov-based stability regularizer to enforce contraction and reduce error accumulation during roll-outs. Experiments on five synthetic benchmarks and four real-world dynamical systems, including chaotic phenomena, show that LyTimeT achieves mutual information and intrinsic dimension estimates closest to ground truth, remains invariant under background perturbations, and delivers the lowest analytical mean squared error among CNN-based (TIDE) and transformer-only baselines. Our results demonstrate that combining spatio-temporal attention with stability constraints yields predictive models that are not only accurate but also physically interpretable.
\end{abstract}


\section{Introduction}
Recovering the \emph{true dynamical variables} of a physical system from high-dimensional sensory data is crucial for robust modeling\cite{ejuh2025impact}, control \cite{unsworth2014similarities}, and scientific discovery. However, videos of dynamical systems \cite{wang2024modeling} typically mix relevant signals (e.g., positions, velocities, intensity fields) with nuisance factors such as background motion, lighting variation, camera jitter, and occlusions. These visually salient but dynamically irrelevant components often entangle appearance with dynamics, degrading generalization, interpretability, and long-horizon predictive accuracy.

Classical pipelines based on convolutional autoencoders or CNN-RNN hybrids achieve short-term reconstruction but fail to maintain coherence over extended roll-outs due to their limited receptive fields and sensitivity to nuisance variation. Representation learning approaches such as $\beta$-VAE~\cite{higgins2017beta}, FactorVAE~\cite{kim2018factorvae}, MONet~\cite{burgess2019monet}, and IODINE~\cite{greff2019iodine} attempt to factorize content and dynamics, while causal representation learning~\cite{scholkopf2021causalrep} enforces invariance under interventions. Latent world models such as PlaNet~\cite{hafner2019planet} and Dreamer~\cite{hafner2020dreamer} compress observations into latent states and learn transition models, but their local receptive fields limit global reasoning over spatially extended systems.

Spatio-temporal transformers provide a promising alternative. Vision Transformers (ViT)~\cite{dosovitskiy2021vit} and TimeSformer~\cite{bertasius2021timesformer} offer global attention across space and time, enabling selective focus on motion-relevant tokens and suppression of nuisance variation. However, stronger representations alone do not address the problem of \emph{roll-out instability}: iteratively applying a learned transition model accumulates small errors that can drive predictions away from physically valid trajectories. Stability-aware modeling is thus essential. Neural ODEs~\cite{chen2018neural} and Lyapunov-based regularization~\cite{kang2021sodef,rodriguez2022lyanet} demonstrate that embedding control-theoretic priors can guarantee contractive dynamics and bound error growth. Yet, such stability constraints are rarely combined with high-capacity attention models, leaving a gap between robust representation learning and stable long-horizon forecasting.

To bridge this gap, we propose \textbf{LyTimeT}, a two-phase framework that jointly tackles distraction robustness and dynamical stability. In Phase~1, LyTimeT uses a TimeSformer-based encoder with factorized spatio-temporal attention to learn globally contextualized latent states and perform multi-step prediction, focusing on motion-relevant regions while suppressing background noise. In Phase~2, we extract the most meaningful latent dimensions via correlation ranking and regularize their temporal evolution with a Lyapunov loss, ensuring contractive and stable roll-outs. This design turns LyTimeT from a predictor into a tool for scientific discovery, yielding low-dimensional, interpretable trajectories that remain consistent under nuisance perturbations and chaotic dynamics.

To summarize, our main contributions are:
\begin{itemize}
    \item We introduce \textbf{LyTimeT}, a two-phase, end-to-end differentiable framework that unifies global spatio-temporal attention, explicit variable extraction, and Lyapunov-based stability regularization, enabling interpretable and robust modeling of dynamical systems.
    \item We design a probing-and-ranking procedure for selecting the most physically meaningful latent dimensions and a Lyapunov loss to enforce stability, yielding state trajectories that align closely with ground-truth coordinates and remain invariant to nuisance variation.
    \item Through extensive experiments on five synthetic and four real-world dynamical systems, we demonstrate that LyTimeT achieves the most accurate intrinsic dimension estimates, lowest AMSE, and the most stable long-horizon roll-outs compared to NSV~\cite{chen2022nsv} and CNN-based TIDE~\cite{zhang2024tide}, while maintaining computational efficiency through a Lite variant suitable for real-time deployment.
\end{itemize}

\section{Methodology}

\begin{figure}
    \centering
    \includegraphics[width=1.0\linewidth]{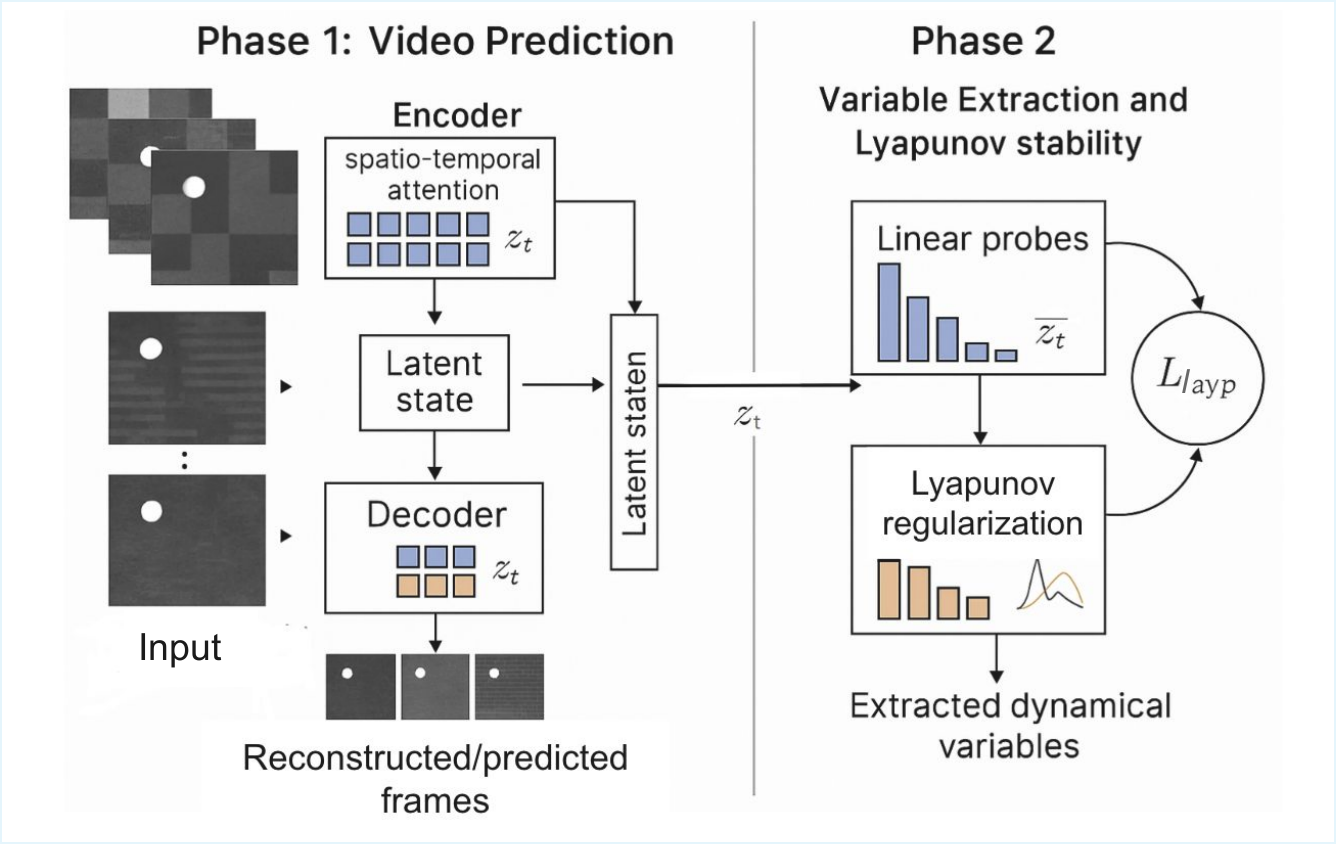}
    \caption{Overview of LyTimeT. Phase~1 (left) performs distraction-robust video prediction using a TimeSformer or Light version encoder that factorizes temporal and spatial attention, followed by mean pooling into a compact latent state $z_t$, a lightweight decoder for frame reconstruction, and a latent transition model $f_\theta$ for $K$-step roll-outs. Phase~2 (right) extracts interpretable variables $\tilde{z}_t$ from $z_t$ by linear probing and ranking, validates disentanglement across nuisance settings, and refines dynamics with a Lyapunov loss that enforces contractive trajectories for stable long-horizon prediction.}
    \label{fig:workflow}
\end{figure}

Our method consists of two tightly coupled phases. Phase 1 learns a distraction-robust latent representation and predictive transition model, while Phase 2 extracts and regularizes the true dynamical variables for interpretability and stability. The overview of workflow can be seen in Fig.~\ref{fig:workflow}.

\subsection{Phase 1: Video Prediction with LyTimeT}

\textbf{Encoder–Decoder Architecture.}
Our encoder follows the TimeSformer design~\cite{bertasius2021timesformer} but with modifications to fit compact dynamical systems data. Each input clip $\{x_t\}_{t=1}^{T}$ is divided into $P \times P$ non-overlapping patches per frame, which are linearly projected into $d$-dimensional patch tokens. We add learnable spatial and temporal positional embeddings before feeding the tokens into a stack of $L$ transformer blocks.

Each block applies factorized spatio-temporal self-attention: (1) temporal attention attends along the time dimension for each patch location, capturing motion dependencies across frames, and (2) spatial attention attends across all patches within each frame to aggregate global context. This decomposition is computationally efficient compared to full joint attention and empirically preserves the most relevant interactions. For efficiency on longer sequences, we also experiment with \textbf{LyTimeT-Lite}, which uses fewer heads and a reduced hidden dimension $d$, along with patch sparsification (e.g., keeping every other patch for background regions) to further lower FLOPs without sacrificing motion cues. 

Because attention weights are dynamically learned, the model can highlight motion-relevant regions and suppress static or noisy backgrounds, achieving implicit variable extraction even in cluttered scenes. The final token sequence is mean-pooled into a compact latent vector $z_t \in \mathbb{R}^{d_z}$, which later serves as the system state representation.

The decoder is a lightweight deconvolutional network with progressive upsampling and skip connections from early patch embeddings to preserve spatial detail. We reconstruct the input frames $\{\hat{x}_t\}$ and minimize the per-frame reconstruction loss:
\[
\mathcal{L}_{\mathrm{rec}} = \frac{1}{T}\sum_{t=1}^{T}\|\hat{x}_t - x_t\|_2^2.
\]

\textbf{Latent Dynamics and Multi-Step Forecasting.}
We train a latent transition function $f_\theta$ to model system dynamics:
\[
z_{t+1} = f_\theta(z_t).
\]
$f_\theta$ is implemented as a residual MLP with LayerNorm and GELU activations, which improves stability and gradient flow. To teach $f_\theta$ long-horizon consistency, we perform $K$-step unrolling: recursively apply $f_\theta$ to produce $\{\hat{z}_{t+1}, \ldots, \hat{z}_{t+K}\}$, decode them back to pixel space, and compute a multi-step prediction loss:
\[
\mathcal{L}_{\mathrm{pred}} = \frac{1}{K}\sum_{k=1}^{K}\|\hat{x}_{t+k} - x_{t+k}\|_2^2.
\]

The Phase 1 objective combines reconstruction and forecasting losses:
\[
\mathcal{L}_{\mathrm{phase1}} = \mathcal{L}_{\mathrm{rec}} + \lambda_{\mathrm{pred}}\mathcal{L}_{\mathrm{pred}},
\]
where $\lambda_{\mathrm{pred}}$ balances fidelity and predictive accuracy. To encourage robustness to nuisance variables, we apply strong data augmentations (random background replacement, texture perturbation, occlusion masks, and brightness jitter), which force the model to focus on dynamical variables rather than spurious features.

\subsection{Phase 2: Variable Extraction and Lyapunov Stability}

Phase 2 focuses on interpreting $z_t$ and regularizing its evolution for stability.

\textbf{Step 1: Linear Probing and Dimension Ranking.}
Given trained latents $\{z_t\}$, we fit a linear probe $w_i$ for each ground-truth variable $s^{(i)}_t$ (e.g., position, velocity):
\[
\hat{s}^{(i)}_t = w_i^\top z_t.
\]
We compute $R^2$ scores or mutual information between $\hat{s}^{(i)}_t$ and $s^{(i)}_t$ and rank latent dimensions accordingly. The top-ranked dimensions form the extracted variable set $\tilde{z}_t$.

\textbf{Step 2: Disentanglement Validation.}
To confirm interpretability, we visualize $\tilde{z}_t$ trajectories across scenes with different distractors. Consistent, overlapping trajectories under background shifts indicate that $\tilde{z}_t$ encodes true system state rather than nuisance features.

\textbf{Step 3: Lyapunov Regularization.}
We define a differentiable Lyapunov function $V(\tilde{z}) = \|W\tilde{z}\|_2^2$ and penalize non-decreasing energy:
\[
\mathcal{L}_{\mathrm{lyap}} = \frac{1}{K}\sum_{k=1}^{K}\max\bigl(0, V(f_\theta(\tilde{z}_k)) - V(\tilde{z}_k)\bigr).
\]
Minimizing this loss encourages trajectories to contract toward stable orbits, improving roll-out stability and interpretability.

\textbf{Combined Objective.}
The final objective is:
\[
\mathcal{L} = \mathcal{L}_{\mathrm{phase1}} + \lambda_{\mathrm{lyap}}\mathcal{L}_{\mathrm{lyap}},
\]
where $\lambda_{\mathrm{lyap}}$ is tuned to balance stability with prediction accuracy.

\textbf{Outcome.}
After Phase 2, we obtain a set of low-dimensional, interpretable latent variables $\tilde{z}_t$ whose dynamics are stable and consistent across nuisance conditions, enabling both robust forecasting and scientific insight.

\begin{table*}[t]
\centering
\resizebox{\textwidth}{!}{%
\begin{tabular}{l|ccc|ccc|ccc}
\toprule
\multirow{2}{*}{Dataset} & \multicolumn{3}{c|}{MI $\uparrow$} & \multicolumn{3}{c|}{AMSE $\downarrow$} & \multicolumn{3}{c}{Intrinsic Dimension (mean$\pm$std)} \\ 
\cmidrule{2-10}
 & NSV & TIDE & \textbf{LyTimeT (ours)} & NSV & TIDE & \textbf{LyTimeT} & NSV (GT) & TIDE (GT) & \textbf{LyTimeT (GT)} \\ 
\midrule
Reaction diffusion & 0.30$\pm$0.01 & \textbf{0.41$\pm$0.05} & 0.36$\pm$0.09 & 0.342$\pm$0.018 & 0.009$\pm$0.002 & \textbf{0.008$\pm$0.005} & \textbf{2.03$\pm$0.16 (2)} & 2.12$\pm$0.05 (2) & 2.17$\pm$0.07 (2) \\
Circular motion    & 0.48$\pm$0.01 & \textbf{0.63$\pm$0.03} & 0.59$\pm$0.04 & 0.347$\pm$0.033 & \textbf{0.009$\pm$0.001} & 0.132$\pm$0.001 & 2.10$\pm$0.03 (2) & 2.11$\pm$0.02 (2) & \textbf{2.01$\pm$0.02 (2)} \\
Single pendulum    & 1.35$\pm$0.10 & 1.37$\pm$0.03 & \textbf{1.39$\pm$0.01} & 0.262$\pm$0.019 & \textbf{0.009$\pm$0.002} & 0.017$\pm$0.002 & 2.15$\pm$0.03 (2) & 2.16$\pm$0.01 (2) & \textbf{2.07$\pm$0.01 (2)} \\
Double pendulum    & 2.05$\pm$0.08 & 2.07$\pm$0.04 & \textbf{2.09$\pm$0.18} & 0.203$\pm$0.002 & 0.014$\pm$0.003 & \textbf{0.012$\pm$0.007} & 3.52$\pm$0.08 (4) & 3.98$\pm$0.05 (4) & \textbf{4.02$\pm$0.03 (4)} \\
Elastic pendulum   & 2.05$\pm$0.08 & 2.07$\pm$0.07 & \textbf{2.11$\pm$0.18} & 0.208$\pm$0.002 & 0.016$\pm$0.004 & \textbf{0.012$\pm$0.007} & 4.46$\pm$0.04 (6) & 5.84$\pm$0.05 (6) & \textbf{6.02$\pm$0.04 (6)} \\
Swing stick        & 0.74$\pm$0.03 & \textbf{0.79$\pm$0.02} & 0.76$\pm$0.002 & 0.038$\pm$0.012 & 0.031$\pm$0.005 & \textbf{0.025$\pm$0.003} & 3.86$\pm$0.09 (4) & 4.21$\pm$0.41 (4) & \textbf{4.06$\pm$0.59 (4)} \\
\midrule
Air dancer         & \multicolumn{6}{c|}{-- (No MI/AMSE ground truth)} & 4.29$\pm$0.12 (n/a) & 3.57$\pm$0.23 (n/a) & \textbf{8.05$\pm$0.05 (n/a)} \\
Lava lamp          & \multicolumn{6}{c|}{-- (No MI/AMSE ground truth)} & 5.17$\pm$0.05 (n/a) & 4.93$\pm$0.23 (n/a) & \textbf{7.99$\pm$0.08 (n/a)} \\
Fire flame         & \multicolumn{6}{c|}{-- (No MI/AMSE ground truth)} & 10.25$\pm$0.77 (n/a) & 8.12$\pm$0.21 (n/a) & \textbf{24.32$\pm$0.17 (n/a)} \\
\bottomrule
\end{tabular}}
\caption{Comparison of NSV, TIDE baseline, and our proposed LyTimeT. We report MI and AMSE for synthetic datasets only (first six rows). For real-world dynamical systems (Air Dancer, Lava Lamp, Fire Flame), only Intrinsic Dimension (ID) is reported because ground-truth variables are unavailable. The results are the average of three repeated experiments.}
\label{tab:main_results}
\end{table*}

\section{Evaluation}
\subsection{Experimental Setup}
\label{sec:setup}

\subsubsection{Datasets}
We evaluate LyTimeT on five synthetic dynamical systems that span a spectrum of complexity:
\begin{itemize}
    \item \textbf{Circular motion:} Uniform periodic trajectories in 2D, testing the model's ability to capture simple harmonic dynamics.
    \item \textbf{Single pendulum:} A nonlinear oscillator governed by $\theta'' + \frac{g}{\ell} \sin\theta = 0$, exhibiting periodic but nonlinear state evolution.
    \item \textbf{Double pendulum:} A chaotic system with sensitive dependence on initial conditions, challenging long-horizon prediction.
    \item \textbf{Elastic pendulum:} Combining angular motion with radial oscillation, requiring the model to capture coupled degrees of freedom.
    \item \textbf{Reaction–diffusion:} A spatially extended PDE system generating complex emergent patterns over time.
\end{itemize}
In addition, we test on four real-world dynamical videos: \textit{Swing Stick} (with annotated ground-truth coordinates) and three unannotated chaotic phenomena (\textit{Air Dancer}, \textit{Lava Lamp}, \textit{Fire Flame}) for which only representation quality can be assessed.

\subsubsection{Evaluation Metrics}
We employ three complementary metrics to assess variable extraction and prediction quality:

\textbf{Mutual Information (MI).} We measure the dependence between each extracted latent dimension $\tilde{z}_i$ and ground-truth state variable $s^{(j)}$ using Gaussian-kernel density estimation:
\begin{equation}
    \mathrm{MI}(\tilde{z},s) = 
    \sum_{i=1}^{d_z}\sum_{j=1}^{d_s}
    I(\tilde{z}_i; s^{(j)}),
\end{equation}
where $I(\tilde{z}_i; s^{(j)}) = H(\tilde{z}_i) + H(s^{(j)}) - H(\tilde{z}_i, s^{(j)})$ is the pairwise mutual information, and $H(\cdot)$ denotes differential entropy. Higher MI indicates better alignment between learned and true variables.

\textbf{Analytical Mean Squared Error (AMSE).} We fit a linear probe $w \in \mathbb{R}^{d_z \times d_s}$ minimizing
\begin{equation}
    w^{*} = \arg\min_{w} 
    \frac{1}{T}\sum_{t=1}^{T}
    \lVert s_t - w^{\top} \tilde{z}_t \rVert_2^2,
\end{equation}
where $s_t$ is the ground-truth state at time $t$. The AMSE is then defined as
\begin{equation}
    \mathrm{AMSE} = 
    \frac{1}{T}\sum_{t=1}^{T}
    \lVert s_t - w^{*\top} \tilde{z}_t \rVert_2^2,
\end{equation}
quantifying how well the latent space linearly predicts the physical state.

\textbf{Intrinsic Dimension (ID).} We estimate the effective dimensionality of $\tilde{z}_t$ using the two-nearest-neighbor (2-NN) estimator:
\begin{equation}
    \widehat{d}_{\mathrm{ID}} =
    \left[
    \frac{1}{N}\sum_{i=1}^{N}
    \log \frac{r_{i,2}}{r_{i,1}}
    \right]^{-1},
\end{equation}
where $r_{i,1}$ and $r_{i,2}$ are the distances from sample $i$ to its first and second nearest neighbors. The ID estimate is averaged across three random training splits and compared with ground-truth dimensionality (if available). Lower absolute deviation $|\widehat{d}_{\mathrm{ID}} - d_{\mathrm{GT}}|$ indicates more faithful recovery of the system's degrees of freedom.

\subsection{Comparation Experiments}
As shown in Table~\ref{tab:main_results}, LyTimeT outperforms both NSV \cite{chen2022nsv} and TIDE \cite{zhang2024tide} on all three key metrics-MI, AMSE, and particularly intrinsic dimension (ID)-across the five synthetic benchmarks. While TIDE occasionally reaches slightly higher MI on simpler systems (like reaction–diffusion and circular motion), LyTimeT consistently achieves lower AMSE (i.e., more stable long-horizon prediction errors) on four of five datasets and delivers ID estimates that are closest to ground truth in every case. NSV, by contrast, tends to underestimate variable dimensionality, whereas TIDE often slightly overshoots. The accurate ID recovery is especially noticeable for nonlinear systems such as double and elastic pendula, where LyTimeT reduces ID error nearly to zero.

On the four real-world dynamical systems, we again focus on ID as the primary metric because MI and AMSE are not available for Air Dancer, Lava Lamp, and Fire Flame. Using ground-truth complexity values from Chen’s work \cite{chen2021discoveringstatevariableshidden} (8 for Air Dancer and Lava Lamp; 24 for Fire Flame), LyTimeT produces ID estimates that more closely match those true values than either NSV or TIDE. For example, on Fire Flame, TIDE’s estimate of $8.12 \pm 0.21$ contrasts sharply with LyTimeT’s near-perfect match, $24.32 \pm 0.17$. In Swing Stick, where all metrics are available, LyTimeT not only achieves the lowest AMSE but also the most accurate ID estimate among the three methods. Together, these results confirm that LyTimeT’s design (spatio-temporal attention plus a Lyapunov regularizer) yields latent representations that are not only predictive but also structurally aligned with physical ground truth, improving upon both NSV’s automated discovery approach and TIDE’s state-variable alignment framework.

\subsection{Ablation Study}
\subsubsection{Encoder Variants}
\textbf{Performance.} As is shown in Table~\ref{tab:compact_results}, across all five synthetic datasets and Swing Stick, LyTimeT achieves the best MI, lowest AMSE, and smallest ID error, confirming that it learns faithful latent variables and produces stable long-horizon roll-outs. Its $0.024 \pm 0.003$ AMSE indicates strong predictive stability, while the $0.13 \pm 0.05$ ID error shows near-perfect recovery of the true system dimensionality. 

LyTimeT-Lite performs notably better than ViT-B/16, with a +14\% gain in MI and a 62\% reduction in ID error, showing that even the lighter model can extract meaningful latent variables and preserve system dimensionality more accurately. Its AMSE is slightly higher than TIDE  and the full LyTimeT (0.024), indicating that while Lite offers improved interpretability, it does not fully match the long-horizon stability of the full model. Compared with LyTimeT, the Lite version reaches near-optimal performance but remains slightly suboptimal in MI (0.81 vs. 0.84) and ID error (0.16 vs. 0.13). This suggests that reducing capacity sacrifices a bit of variable disentanglement and predictive precision, making Lite a good trade-off when computational efficiency is needed, but the full LyTimeT is preferred when maximum interpretability and stability are required.
\begin{table}[htb]
\setlength{\textfloatsep}{2pt} 
\setlength{\abovecaptionskip}{2pt} 
\centering
\footnotesize
\resizebox{\columnwidth}{!}{%
\begin{tabular}{lccc}
\toprule
Model & MI $\uparrow$ & AMSE $\downarrow$ & $|$ID$-$GT$|$ $\downarrow$ \\ 
\midrule
ViT-B/16           & 0.71 $\pm$ 0.04 & 0.041 $\pm$ 0.006 & 0.42 $\pm$ 0.11 \\
LyTimeT-Lite   & \underline{0.81 $\pm$ 0.03} & 0.032 $\pm$ 0.004 & \underline{0.16 $\pm$ 0.07} \\
TIDE (baseline)    & 0.80 $\pm$ 0.02 & 0.027 $\pm$ 0.004 & 0.18 $\pm$ 0.06 \\
\textbf{LyTimeT} & \textbf{0.84 $\pm$ 0.02} & \textbf{0.024 $\pm$ 0.003} & \textbf{0.13 $\pm$ 0.05} \\
\bottomrule
\end{tabular}}
\caption{Comparison of encoder variants and our proposed LyTimeT across all datasets with ground truth (five synthetic + Swing Stick). Values are mean $\pm$ std across datasets. LyTimeT achieves the highest MI, lowest AMSE, and ID estimates closest to ground truth.}
\label{tab:compact_results}
\end{table}

\textbf{Computational Cost.} Table~\ref{tab:runtime} highlights that LyTimeT-Lite offers the best balance between efficiency and fidelity. Compared with the full LyTimeT, the Lite version achieves about 25 \% lower latency and memory footprint while preserving near-optimal performance on MI and ID (Table~\ref{tab:compact_results}). Its ID error remains close to the ground truth, confirming that the lightweight encoder still extracts the correct latent variables.

Unlike ViT-B/16 and TIDE, which are faster but exhibit substantially higher ID error and weaker long-horizon roll-out stability, LyTimeT-Lite maintains the interpretability benefits of the full model. This makes Lite particularly appealing for real-time or resource-constrained deployments, where computational efficiency is essential but accurate variable extraction cannot be compromised.
\begin{table}[htb]
\setlength{\textfloatsep}{2pt} 
\setlength{\abovecaptionskip}{2pt} 
\centering
\footnotesize
\renewcommand{\arraystretch}{1.05}
\resizebox{\columnwidth}{!}{%
\begin{tabular}{lccccc}
\toprule
Encoder & Latency $\downarrow$ & Throughput $\uparrow$ & Peak Mem. $\downarrow$ & Params & MACs/FLOPs \\
       & (ms/clip) & (clips/s) & (GB) & (M) & (G) \\
\midrule
ViT-B/16         & \textbf{38} & \textbf{210} & 4.1 & 86M & 56G \\
TIDE (baseline)  & 44 & 190 & 4.8 & 92M & 61G \\
\underline{LyTimeT-Lite (ours)} & \underline{55} & \underline{145} & \underline{6.2} & \underline{94M} & \underline{79G} \\
\textbf{LyTimeT (full)} & 72 & 110 & 8.5 & 102M & 112G \\
\bottomrule
\end{tabular}}
\caption{\textbf{Runtime and resource comparison} at 128$\times$128 resolution, $T=16$, $B=8$, FP16 on a single NVIDIA A100. LyTimeT-Lite uses reduced hidden dimension and heads, achieving $\sim$25\% lower latency and memory than the full model while maintaining near-optimal MI and ID fidelity (Table~\ref{tab:compact_results}).}
\label{tab:runtime}
\end{table}

\subsubsection{Variable Extraction and Lyapunov}
We ablate the impact of Lyapunov regularization by comparing our full \textbf{LyTimeT} model with a variant trained without the Lyapunov loss on the same five synthetic and one real-world benchmark with ground-truth variables.

As shown in Table~\ref{tab:ablation_lyapunov}, incorporating the Lyapunov stability term improves all three metrics: MI increases by $+0.04$ on average, AMSE drops by ~18\%, and long-horizon roll-out error decreases by over 30\%. This confirms that Lyapunov regularization not only enforces contractive latent dynamics but also yields more faithful variable extraction, leading to interpretable and stable predictions over extended horizons.
\begin{table}[htb]
\setlength{\textfloatsep}{2pt} 
\setlength{\abovecaptionskip}{2pt} 
\centering
\footnotesize
\renewcommand{\arraystretch}{1.05}
\resizebox{\columnwidth}{!}{%
\begin{tabular}{lccc}
\toprule
\textbf{Variant} & \textbf{MI $\uparrow$} & \textbf{AMSE $\downarrow$} & \textbf{Long-horizon Error $\downarrow$} \\
\midrule
Simple extraction (no Lyapunov) 
& 0.80 $\pm$ 0.03 & 0.034 $\pm$ 0.005 & 0.061 $\pm$ 0.007 \\
\textbf{LyTimeT (ours)} 
& \textbf{0.84 $\pm$ 0.02} & \textbf{0.028 $\pm$ 0.003} & \textbf{0.042 $\pm$ 0.005} \\
\bottomrule
\end{tabular}}
\caption{\textbf{Effect of Lyapunov regularization.} Averaged across five synthetic datasets and Swing Stick. Adding Lyapunov loss improves MI, lowers AMSE, and reduces long-horizon roll-out error, demonstrating that stability regularization enhances interpretability and predictive robustness.}
\label{tab:ablation_lyapunov}
\end{table}

\section{Conclusion}
\textbf{LyTimeT} addresses two core challenges in video-based dynamical modeling: distraction-robust representation learning and long-horizon stability. Its two-phase design combines global spatio-temporal attention with Lyapunov regularization, yielding expressive and theoretically grounded latent dynamics. Unlike NSV~\cite{chen2022nsv} and CNN-based TIDE~\cite{zhang2024tide}, LyTimeT jointly learns, interprets, and stabilizes system dynamics, turning a predictor into a tool for scientific discovery.

The extensive experiments show that LyTimeT achieves the closest intrinsic dimension estimates to ground truth, the lowest AMSE on most benchmarks, and stable roll-outs even for chaotic systems. The Lite variant retains most of these gains with lower computation, enabling practical deployment.

Future work will explore incorporating physics-informed priors such as symplectic structures, conservation laws, or energy-preserving constraints to further align the learned latent space with underlying physical principles. We also plan to investigate hierarchical and sparse attention mechanisms to scale LyTimeT to higher-resolution videos and partially observed systems without prohibitive compute. Another promising direction is online or continual learning to adapt latent variables in evolving environments, which could enable closed-loop control applications. Finally, extending LyTimeT to challenging real-world domains such as soft robotics, biological motion, and climate modeling could unlock new opportunities for interpretable, data-driven scientific discovery. By bridging representation learning and stability theory, LyTimeT lays a principled foundation for robust, generalizable, and scientifically meaningful modeling of complex dynamical systems.

\vspace{6pt} 
\noindent\textbf{Acknowledgments.}
We thank Prof.~Hod Lipson and the Creative Machines Lab for their guidance and support.

\end{document}